\documentclass[letterpaper, 10 pt, conference]{ieeeconf}  %

\IEEEoverridecommandlockouts                              %

\usepackage{graphics} %
\usepackage{epsfig} %
\usepackage{mathptmx} %
\usepackage{times} %
\usepackage{amsmath} %
\usepackage{amssymb}  %

\usepackage{rotating}
\usepackage{array}
\usepackage[colorlinks]{hyperref}
\usepackage{placeins}
\usepackage{animate}
\usepackage{amsmath}
\usepackage{stfloats}   %

\usepackage{amsfonts}
\usepackage{amssymb}
\usepackage{comment}
\usepackage{float}
\usepackage{caption}
\usepackage{color}
\usepackage{algorithmic}
\usepackage{algorithm}
\usepackage{subfigure}
\usepackage{multirow}
\usepackage{booktabs}
\usepackage{subfigure}

\title{\LARGE \bf
Modeling of AUV Dynamics with Limited Resources: Efficient Online Learning Using Uncertainty}

\author{Michal Te\v{s}nar$^{1}$, Bilal Wehbe$^{2}$, Matias Valdenegro-Toro$^{1}$%
\thanks{$^{1}$Department of AI, Bernoulli Institute, University of Groningen, The Netherlands
        {\tt\small m.a.valdenegro.toro@rug.nl}}%
\thanks{$^{2}$Robotics Innovation Center, German Research Center for Artificial Intelligence, Bremen, Germany.
        {\tt\small bilal.wehbe@dfki.de}}%
}

\begin{document}

\maketitle
\thispagestyle{empty}
\pagestyle{empty}

\begin{abstract} Machine learning proves effective in constructing dynamics models from data, especially for underwater vehicles. Continuous refinement of these models using incoming data streams, however, often requires storage of an overwhelming amount of redundant data. This work investigates the use of uncertainty in the selection of data points to rehearse in online learning when storage capacity is constrained. The models are learned using an ensemble of multilayer perceptrons as they perform well at predicting epistemic uncertainty. We present three novel approaches: the \textit{Threshold} method, which excludes samples with uncertainty below a specified threshold, the \textit{Greedy} method, designed to maximize uncertainty among the stored points, and \textit{Threshold-Greedy}, which combines the previous two approaches. The methods are assessed on data collected by an underwater vehicle Dagon. Comparison with baselines reveals that the \textit{Threshold} exhibits enhanced stability throughout the learning process and also yields a model with the least cumulative testing loss. We also conducted detailed analyses on the impact of model parameters and storage size on the performance of the models, as well as a comparison of three different uncertainty estimation methods.
\end{abstract}

\section{INTRODUCTION}
In recent years, machine learning techniques have become popular for obtaining information from streams of data, thereby replacing manual data aggregation. In robotics, machine learning can be applied to learn a dynamics model of the controlled vehicle, forming basis for control, localization and simulation frameworks. The dynamics model describes the behavior of the vehicle over time, based on its states and inputs. Instead of constructing the equations that govern the model, the relation can be inferred from the states of the vehicle that were recorded during its operation \cite{wehbe2019framework}. This can be done by \textit{direct modeling} which aims to create a model that predicts the next state given the current state of the model \cite{thuruthel2017learning}. In underwater robotics, this allows us to directly learn complex models without having to sacrifice performance due to simplifying assumptions while avoiding expensive computations of analysis of \textit{Navier-Stokes} equations that describe the motion of fluids. Due to the non-linearity of the task, general function approximation methods such as neural networks have shown higher fidelity compared to finite-dimensional approximations of the system's hydrodynamics \cite{wehbe2017learning}. The obtained model can later be applied in the control of \textit{autonomous underwater vehicles} (AUVs), for example, in model predictive control, which uses the dynamics model to predict the future states of the vehicle to find the most optimal action at each time to achieve its goal \cite{henson1998nonlinear}.

Furthermore, to achieve optimal control in a changing environment, it is desirable to be able to continuously update the dynamics model of the vehicle during its operation based on the newly collected data. The vehicle might perform new actions or encounter new environment configurations, which might not yet be included in the model. Furthermore, there can be changes in the environment, such as changes in the density of the surrounding liquid, and in the robot itself as the robot's body can wear over time or the thrusters break down \cite{wehbe2020long}. At the same time, however, the model needs to preserve its past knowledge. Simply retraining the learner on incoming data might lead to loss of previously obtained knowledge, commonly referred to as \textit{catastrophic forgetting} \cite{french1999catastrophic}. This can be combatted by preserving points seen in the past and retraining on them. This is usually referred to as \textit{rehearsal}, as the model learns on samples it has already seen \cite{verwimp2021rehearsal}. Overall, the goal of this effort is commonly referred to as \textit{incremental online learning}, the process of adapting the model without losing previously learned information \cite{8329822}.

\begin{figure}[t]
    \centering
    \includegraphics[width=\linewidth]{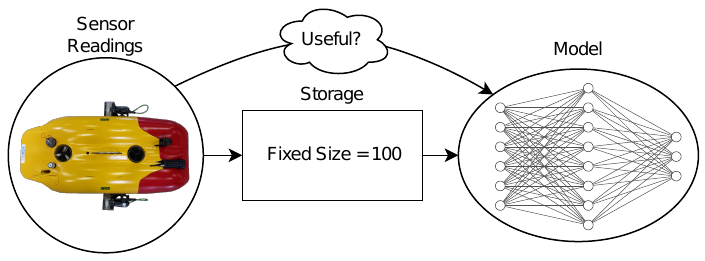}
    \caption{Problem addressed in this work. \textmd{Gathered data contains redundancy. To restrict the size of storage we determine whether points are useful for learning by quantifying uncertainty.}}
    \label{fig:problem-statement}
\end{figure}

One of the possible solutions for the stated problem would be to continuously store all collected data and retrain the learner from scratch. However, this is not feasible in real-life scenarios due to the limited resources of AUVs. What is more, many of the collected data points inevitably hold redundancy, as the AUV might be performing the same maneuver repeatedly. This gives motivation for the selection of the most informative points to store and train on. This is visualized in Figure \ref{fig:problem-statement}.

One can use the concept of uncertainty to assess the collected samples. Intuitively, uncertainty gives us useful information about the sample: the samples the learner is the most uncertain about might contain the most useful information to be learned. This concept is also known as \textit{active learning}: models can achieve greater accuracy on smaller amounts of data if allowed to choose which data to train on \cite{settles2009active}. We can equip our learners with an uncertainty quantification method to be able to predict the uncertainty of the predictions. We are interested in the \textit{epistemic} uncertainty, which quantifies the uncertainty of the prediction given knowledge in the model \cite{valdenegro2022deeper}. In regression settings, the uncertainty of a model can be interpreted as the variance of a model around a certain prediction mean, or as a confidence interval of the prediction.

In our task, knowledge of epistemic uncertainty can be exploited, as we can use it as a determining criterion for whether a point is informative for the model. We can decide whether the sample is worth storing and training on if the model is uncertain about it enough. The uncertainty estimation can serve also other purposes. It can be used to give meaning to the prediction of the output of the model in deployment. In control, it can be used as an indicator of the confidence of the vehicle in the current manipulation scenario. In pose estimation, it can be used in the Kalman filter, which can benefit from the knowledge of uncertainty of the model dynamics \cite{cantelobre2020real}. %

The contribution of this work lies in the investigation of active learning in complex regression settings. More specifically, we want to show how uncertainty can be exploited to increase efficiency and stability in online incremental learning of AUV dynamics regression tasks while resources are constrained. We want to evaluate if by using uncertainty, we can reduce the amount of storage and training needed to perform online learning of the AUV dynamics model. In this manuscript, (1) we provide three techniques for performing online active learning using uncertainty estimation, namely \emph{Greedy} method, \emph{Threshold} method, and \emph{Threshold-Greedy}, (2) we compare the performance of proposed methods against online leaning baselines such as first-in-first-out (FIFO), first-in-random-out (FIRO)  and random-in-random-out (RIRO) using real experimental data collected with the AUV Dagon (see Fig.~\ref{fig:problem-statement}). Furthermore, (3) we compare the performance of three different uncertainty estimation techniques for neural networks, namely, Ensembles, Monte-Carlo Dropout, and Flipout methods.

\section{Related Works}\label{sec:related-works}

Learning dynamics and kinematic models is certainly not a new task. The fact that learning a model from collected sensor data is more feasible than explicitly defining the model has been long recognized and heavily researched in many different scenarios. Two paradigms are commonly discussed in the literature for predicting motion dynamics: (1) identifying the coefficients of an established model \cite{paine2018adaptive,gibson2018hydrodynamic,harris2023stable}, and (2) employing machine learning-based function approximation methods \cite{wehbe2019framework,ramirez2021dynamic,bande2021online}. For supervised learning, labeled data is usually expensive, which gives incentive to investigate the strategy of active learning in this context \cite{cohn1996active, dasgupta2004analysis}.

Generally, there are two types of learning, first is a global approach, which tries to approximate one function across the whole dataset. The popular choices for this include \textit{gaussian regression process} \cite{williams1995gaussian}, \textit{support vector machine regression} \cite{scholkopf2000new}, and \textit{variational Bayes for mixture models} \cite{ghahramani1999variational}. Conversely, in an incremental local approach, one can incrementally fit the data locally by the multitude of functions \cite{vijayakumar2005incremental}. In this work, we will focus on the global methods. Even though they are harder to tune in scenarios where little prior knowledge is available about the complexity of the dataset, their main advantage is that they allow us to create an uncertainty measurement on a model. For that, the model needs to be consistent over the whole input space. Furthermore, the conclusion about global methods can be later used in other areas of research, where global models are the current state of the art. 

The paradigm of incremental online active learning has not yet been heavily explored in regression settings. It has been widely explored in the realm of classification, to tackle problems such as concept drift \cite{sculley2007online, he2020incremental}. In regression settings, it has been explored in a simple setting of linear regression \cite{chen2022online}. However, this work will expand this approach to a more complex regression task. Identification of a dynamics model of AUV is highly non-linear and many dimensional. This gives us the opportunity to explore the applicability of these methods in more complex settings. Furthermore, the methods will be tested on real-world collected data, which will show how the methods perform with natural aleatoric certainty generated by the noise in the sensors.

We observe that the problem of updating the knowledge if the model gets outdated is called \textit{adaptive learning}, that is if we have concept drift in our case, we might change the model, we could be certain but we still need to re-learn it \cite{loeffel2017adaptive}. We decide to ignore this issue, as it is outside of the scope of this work.

In the field of robot learning in particular, there is already a track history of investigation of learning models. In \cite{paine2023ensemble}, three methods for online system identification are compared: recurrent neural network, adaptive identification (AID), and recursive least squares method. Showing that AID achieved the lowest MAE during online training. long-short term memory networks (LSTMs) are also popular for modeling model vehicle dynamics. In \cite{bande2021online}, LSTMs in combination with a memory-efficient rehearsal method were used for learning AUV dynamics. This can help replace expensive sensors, as shown in \cite{topini2020lstm}, where LSTMs are used to learn surge-sway velocities of AUV, avoiding the need for Doppler Velocity Log. Furthermore, \cite{wang2021efficient} uses Gaussian model approach to update an end-to-end data-driven model of vehicle dynamics of driverless cars.

\section{Proposed Method}\label{sec:proposed-methods}
To reach our goal, we need to implement uncertainty into the selection of data in online learning and compare the performance of that model to a baseline model that does not use this technique. To do this, we will first describe the specification of the learning task that we are tackling and introduce the dataset. Then we will specify in detail how the incremental online learning approaches, both baselines and uncertainty-augmented methods. Then we will discuss the evaluation criteria and machine learning background, including uncertainty quantification.

\subsection{Model Learning}

The aim is to learn a data-driven dynamics model. We will do this by direct modeling: directly relating the inputs and the outputs. As this work is an extension of \cite{wehbe2017learning}, this section will closely follow the derivations given in section \textit{II.} and \textit{III.} of the paper. The motion vehicles in three-dimensional spaces can be described using 6 degrees of freedom. The forward motion is referred to as \textit{surge}, sideways motion is \textit{sway}, and upwards and downwards \textit{heave}. Additionally, to describe the rotation of the body in space, which is for direction side-to-side \textit{roll}, up-down \textit{pitch}, and for left-right \textit{yaw}. To obtain the vehicle dynamics, we can record velocities and acceleration in those directions. Those can be integrated to obtain the trajectory of the vehicle.

\begin{figure}[t]
    \centering
    \includegraphics[width=\linewidth]{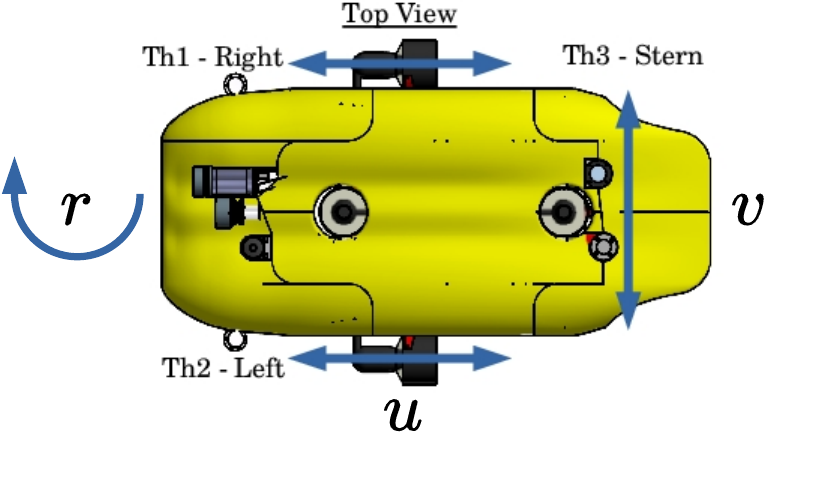}
    \caption{Top view of the Dagon AUV, its degrees of freedom and related thrusters. \textmd{The vehicle was fixed in the horizontal plane. It was controlled in surge, sway, and yaw ($u, v, r$) and steered by 3 thrusters $(n_1, n_2, n_3)$. Figure adapted from \cite{wehbe2017learning}.}}
    \label{fig:dagon-setup}
\end{figure}

\subsection{Online Learning Methods}

\begin{figure}[t]
    \centering
    \includegraphics[width=\linewidth]{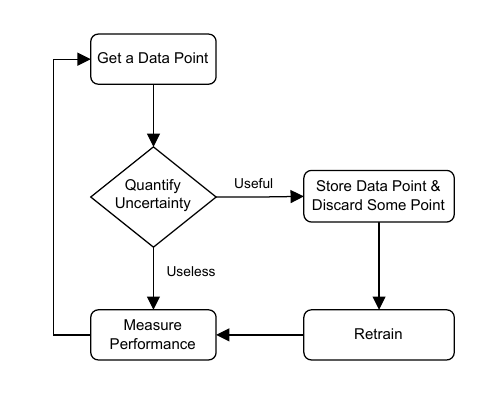}
    \caption{Diagram of applied method. \textmd{The online learning process can be augmented with uncertainty quantification. Based on uncertainty we can determine the usefulness of a point. Using it, we can choose to store and train on the incoming point, or whether to skip it.}}
    \label{fig:learning-process}
\end{figure}

In this section, we explain how we can learn from data online in an incremental fashion. We will start by describing baseline approaches and then we will follow with the techniques that exploit uncertainty. For each of the models, we assume an incremental online learning process. This means that the model gets new data points served one at a time and can learn from them and store them, or it can reject them as visualized in Figure \ref{fig:learning-process}. The process of selection of points for each invididual selection technique is visualized in Figure \ref{gif:all_figs}.

\subsubsection{Baseline Approaches}

\textbf{Offline}: We assume all data was already collected in the past and we train on all of them. The performance of this model can be assumed to be the optimal one. This can be assumed to be the optimal solution.

\noindent \textbf{First-In-First-Out (FIFO)}: In each iteration, the model will forget the data point it has seen the latest and add the newest collected point. We expect this method to perform poorly due to catastrophic forgetting, the model will forget previously learned information as it will fit on the most recent data. This is visualized in Figure \ref{gif:fifo}.

\noindent \textbf{First-In-Random-Out (FIRO)}: To have the model maintain a more evenly spread distribution over the dataset, we consider instead forgetting a random point from the currently stored set. We expect this model to perform better than FIFO since it will preserve some of its knowledge. This is visualized in Figure \ref{gif:firo}.

\noindent \textbf{Random-In-Random-Out (RIRO)}: This next approach will choose to learn and store an incoming point with probability $p$, that will need to be tuned to simulate similar effect to skipping points in uncertainty-based methods. This model should obtain a more even distribution over the dataset. With smaller $p$ should take longer to establish this distribution, but remain more stable later on. This is visualized in Figure \ref{gif:riro}.

\subsubsection{Approaches with Uncertainty Quantification}

For each point, the model can run to determine its uncertainty on this point. This will be exploited to decide whether to include the point in the dataset or not. The model will be retrained each time that a new point is added to the dataset.

\noindent \textbf{Greedy}: This method greedily selects the most interesting points to learn from, that is the points that are the most uncertain. When the incoming point is assessed for uncertainty, the same is done for the currently stored points. If the incoming point has higher uncertainty than any point in the current set, it will be substituted for this point. This is visualized in Figure \ref{gif:greedy}.

\noindent \textbf{Threshold}: To avoid re-learning redundant information, this model chooses to avoid points whose uncertainty is low. The uncertainty must be above a bound $t$, which is a hyperparameter to be determined. The point chosen to be discarded from the currently held dataset is selected randomly. This is visualized in Figure \ref{gif:threshold}.

\noindent \textbf{Threshold-Greedy}: This approach aims to combine the two previously presented techniques. Each point will be first compared against the uncertainty threshold $t$. Subsequently, if it needs to be inserted into the dataset, the least informative sample of the dataset will be discarded. This is visualized in Figure \ref{gif:threshold-greedy}.

We expect that the methods that utilize uncertainty to outperform the rest. To train a good model, it might be logical to just distribute the data evenly over the input space. However, it might be, that some parts of the space are very non-linear, and therefore more difficult to learn. Those might benefit from having more stored data points. This could be even better captured by the uncertainty, as the model will remain more uncertain in those regions.

\section{Experimental Setup}\label{sec:experiment-setup}

Firstly, we will fix the parameter $p$ in RIRO and $t$ in Threshold and Threshold-Greedy. Then we will perform hyperparameter tuning of all model machine learning parameters. Then we will separately investigate the influence of $t$ and $p$ on the learning process. We will pick the most optimal parameters for each method and we will compare the models once again, to reach a better conclusion of the comparative performance. Lastly, we will investigate the size of the buffer of the models and see how their performance changes with respect to this parameter. %

\subsection{Dataset}

The data that we will be using was collected by the autonomous underwater vehicle Dagon. Dagon is a vehicle specifically designed for scientific testing and evaluation of algorithms, especially in underwater visual mapping and surveying near-shore continental shelves. More information can be found online \cite{hildebrandt2010design}.

\begin{figure}[t]
    \centering
    \includegraphics[width=\linewidth]{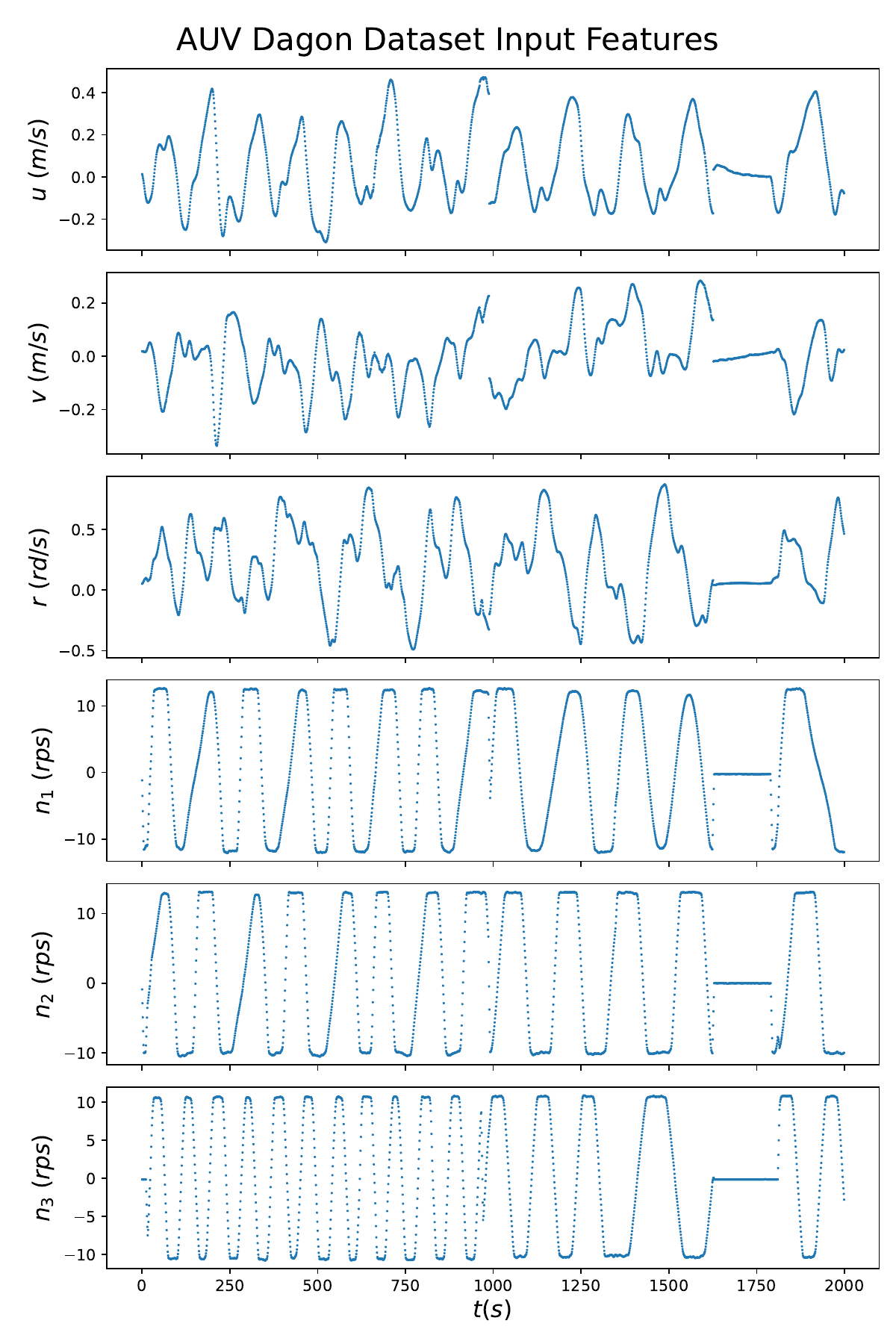}
    \caption{Input features of the Dagon dataset}
    \label{fig:dagon-input-features}
\end{figure}

\begin{figure}[t]
    \centering
    \includegraphics[width=\linewidth]{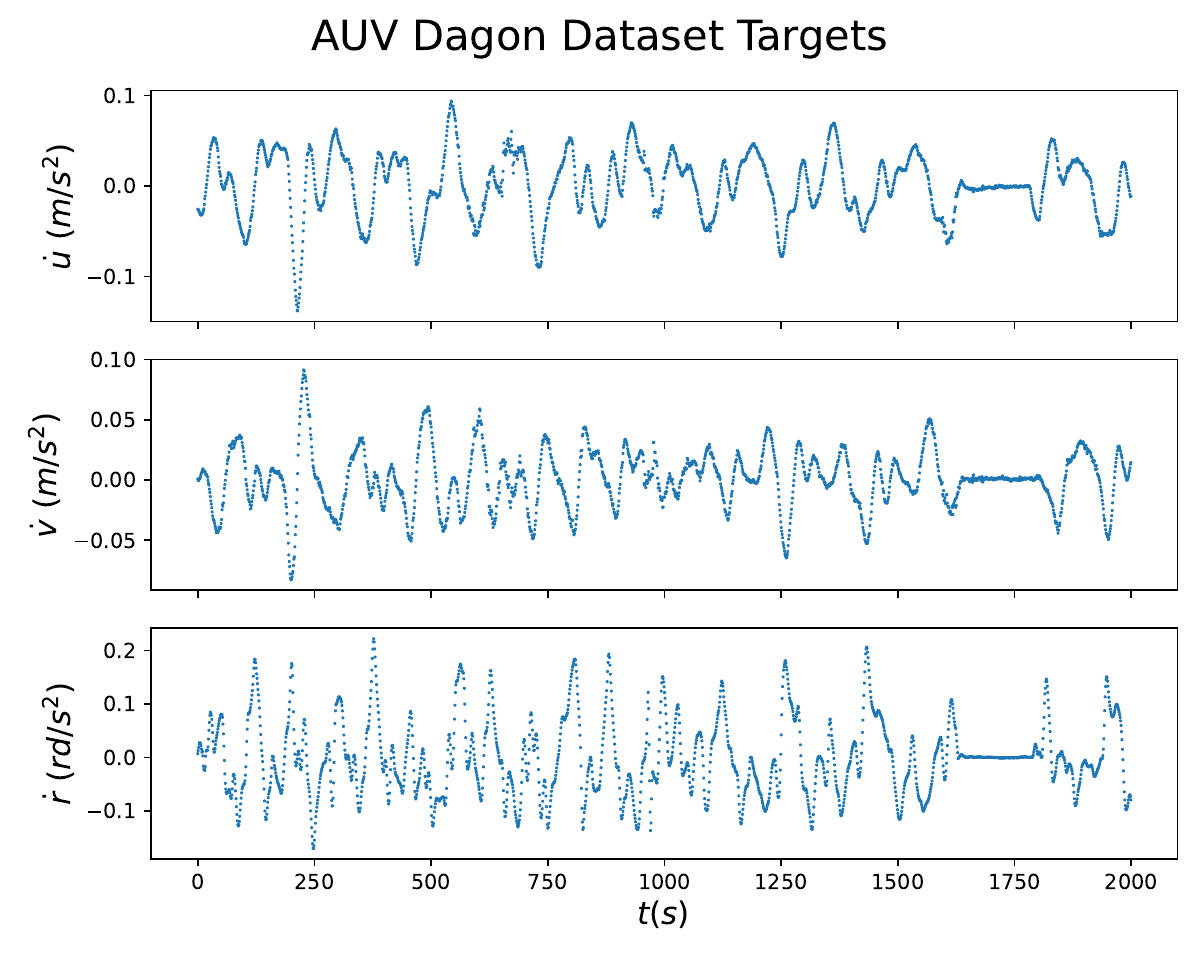}
    \caption{Targets of the Dagon dataset}
    \label{fig:dagon-targets}
\end{figure}

To collect the dataset, the vehicle was driven in the salty water basin of the research center. The vehicle was stabilized to drive in the horizontal plane being manipulated by 3 thrusters. The thrusters were given sinusoidal input with different periods. During the experiment, the linear and angular velocities of the vehicle were captured. This data was numerically differentiated to obtain acceleration \cite{wehbe2017learning}. The first 2000 features and targets of the datasets are visualized in Figures \ref{fig:dagon-input-features} and \ref{fig:dagon-targets}.

Following the notation from \cite{wehbe2017learning}, we can denote the first derivatives of the surge, sway, heave, roll, pitch, and yaw as $\nu =(u, v, w, p, q, r) \in \mathbb{R}^{6}$ and the inputs of the 5 thrusters as $n = (n_1, n_2, n_3, n_4, n_5)$. Given the above-described restrictions, we are learning function $\mathcal{F}$ described as:
\begin{equation}
(\dot{u}, \dot{v}, \dot{r}) =\mathcal{F}(u, v, r, n_1, n_2, n_3)
\end{equation}

To assess the performance of the model, a test set was withheld before the training process. As the data from the robot was collected over time, the test set was sampled at evenly spaced intervals throughout the dataset set. The dataset split follows the ratio 60/20/20\% for training, validation, and testing sets.

\subsection{Metrics}

To measure the instantaneous performance of the model, we will use MSE and $R^2$. We note that for multivariable regression, the $R^2$ cannot clearly be defined, therefore we will consider the mean $R^2$ of all vector components.

To summarize the performance of the model over the learning process, we will use the cumulative MSE. We will also look at the development of the predicted uncertainty of the model on the incoming points. Lastly, as a measure of saved resources, we will count the number of skipped points.

\subsection{Uncertainty Quantification}

The use of uncertainty builds on the assumption, that the points with higher uncertainty possess information that is more interesting to the learner. For our tasks, we need to estimate epistemic uncertainty, which estimates how the model is uncertain on the input data. There are many different uncertainty estimation techniques, however, to estimate epistemic uncertainty, ensembles perform the best \cite{valdenegro2022deeper, valdenegro2021exploring}. For comparison, we also repeat the experiments using Monte Carlo Dropout \cite{gal2016dropout} and Flipout \cite{wen2018flipout}.

\subsection{Model}

The network architecture used in the experiments is a multilayer perceptron. The model uses the ReLU activation function in hidden layers and linear activation in the last layer. The model is trained using the Adam optimizer with MSE loss. The number of hidden layers (\{1, 2, 3, 4\}), units in each hidden layer (\{4, 8, 16, 32, 64\}), learning rate (\{$10^{-2}, 10^{-3}, 10^{-4}, 10^{-5}$\}) and batch size (\{1, 2, 4, 8, 16\}) were tuned using random search of 60 iterations for each of the methods separately. Additionally, the parameter of patience (\{3, 5, 9\}) in the early stopping criterion was tuned. The uncertainty estimation methods were implemented into this architecture.

\subsubsection{Ensembles}

This method of ensembles \cite{lakshminarayanan2017simple} relies on the simple, yet powerful idea of combining multiple models of the same kind to obtain output uncertainty. In our experiment, we consider a simple ensemble of 10 independent models. The mean prediction of the models is then output. The standard deviation of the model is a measure of uncertainty.

Note that this ensemble technique is different from the so-called deep ensembles. The deep ensemble uses a model with two heads, one that predicts mean and one that predicts variance. This allows us to estimate both aleatoric and epistemic uncertainty. The head that predicts the mean, is supervised by the label data and the variance with the Gaussian negative-log-likelihood \cite{lakshminarayanan2017simple}. However, as our goal is to only estimate epistemic uncertainty, a simple ensemble is enough.

\subsubsection{Monte Carlo Dropout}

Another method to estimate uncertainty is to use Monte Carlo Dropout. This method is based on the idea that the dropout layer in the neural network can be used at inference time. The dropout layer is a layer that randomly sets some of the neurons to zero. This can be used to estimate uncertainty, by taking multiple forward passes, on which we can assess the mean and the variance of the prediction \cite{gal2016dropout}. In our experiment, we use a dropout probability of 0.2 between the layers. The prediction uses 10 forward passes.

\subsubsection{Flipout}

Flipout is a method that is based on the idea of Monte Carlo Dropout, however, it is more efficient. It is based on the idea that of sampling pseudo-independent weight perturbations for samples at inference \cite{wen2018flipout}. In our setup, we utilize one Flipout layer as the last layer of the network. To obtain uncertainty, we take 10 samples on prediction.

\subsection{Learning Setup}

To run the experiments, we used a simple ensemble with 10 estimators. To account for the differences in training requirements throughout the incremental online learning process, the model has the maximum amount of epochs it can learn on set to 100, however, each strategy can use its tuned patience parameter. Each of the models learns using MSE loss, with Adam, which is built in Keras. The number of layers, the units in each layer, the learning rate, batch size, and patience were tuned independently.

The experiments were implemented in Python 3 using the machine learning framework Keras, the repository is available publicly\footnote{https://github.com/MichalTesnar/mystery}. To estimate uncertainty, the library \emph{Keras Uncertainty}\footnote{https://github.com/mvaldenegro/keras-uncertainty} was used. It provides all necessary utilities for estimating uncertainty on multi-layer perception. To perform the resource-demanding experiments, the high-performance cluster Habrók of the University of Groningen was used.
\section{Results}\label{sec:results}

To find whether uncertainty-based selection can help us decrease the size of the dataset stored in the buffer, we compared the performance of different models on the above-explained task. We first walk through the results for simple ensembles, and then we compare the exact same experiments with Flipout and Dropout. For Simple Ensembles, we first fitted the parameters according to a unified procedure. The values of $p$ in RIRO and $t$ in Threshold and Threshold greedy were fixed with arbitrary values, so next up we investigate the influence of those. Subsequently, we will present a final comparison of results with optimal parameters. Using those optimal parameters, we investigate the role of the size of the buffer on the performance of the models. Lastly, we will present the results of the experiments with Flipout and Dropout.

\subsection{Parameter \texorpdfstring{$p, t$}{Lg} Study}

\begin{figure}[t]
    \centering
    \includegraphics[width=0.46\linewidth]{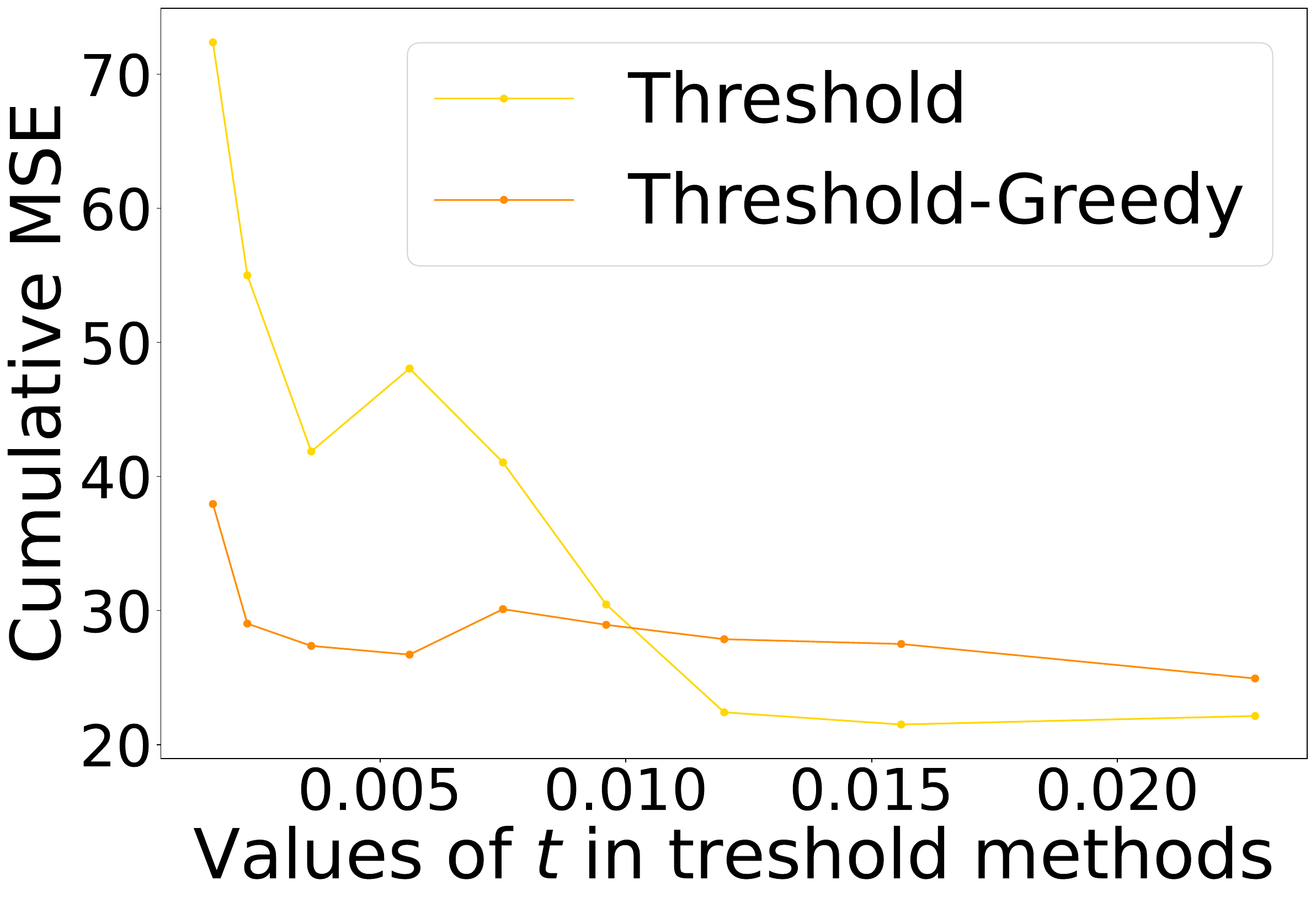}
    \includegraphics[width=0.46\linewidth]{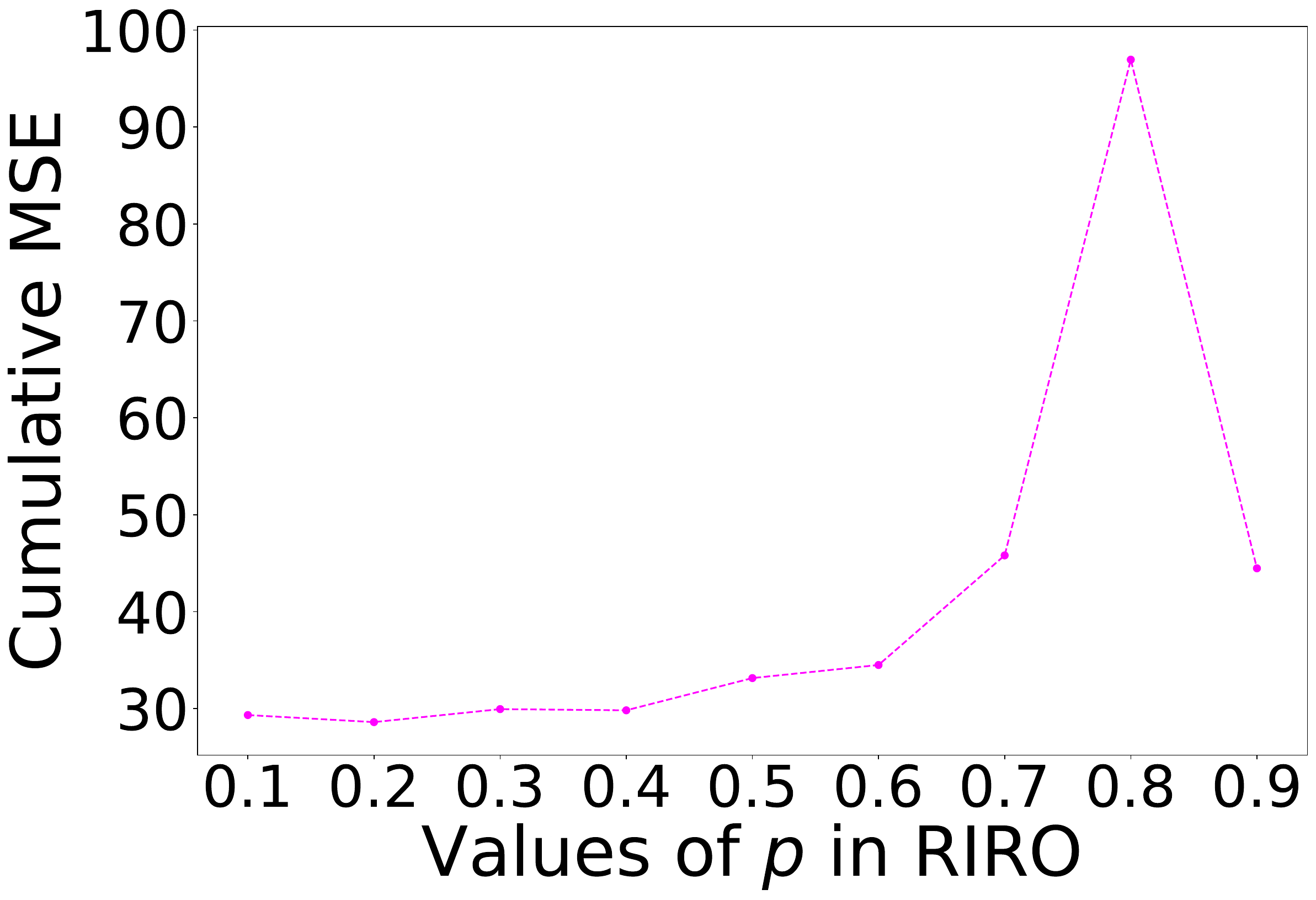}
    \caption{Tuning of Extra Parameters. The left figure shows the cumulative MSE for Threshold and Threshold-Greedy models with varying parameter $t$. The right figure shows the cumulative MSE for RIRO models with varying parameter $p$. In both cases, the y-axis denotes the final cumulative MSE after the learning process.}
    \label{fig:Extra_Parameters_Tuning}
\end{figure}

We investigated the effect of the parameter $p$ in RIRO. The parameter $p$ represents the probability of accepting a point into the dataset. We evenly sampled the space of possibilities, taking $p$ from the set $\{0.1, 0.2, 0.3, 0.4, 0.5, 0.6, 0.7, 0.8, 0.9\}$. Similarly, the parameter $t$ was tested in Threshold and Threshold-Greedy. To sample it, we divided uncertainty into percentiles based on the uncertainty predictions of baselines, which use all the points (FIFO and FIRO). We recorded the uncertainty prediction of all points in the training set and divided the points into uncertainty level percentile. We did this both for FIFO and FIRO and took the mean of the percentiles, sampling 9 evenly spaced values from those. For each of these values of $t$, we performed the experiments once again. The final cumulative MSE as a function of $p$ and $t$ can be seen in Figure \ref{fig:Extra_Parameters_Tuning}.

\subsection{Optimal Parameters}

\begin{figure*}[t]
    \centering \includegraphics[width=\linewidth]{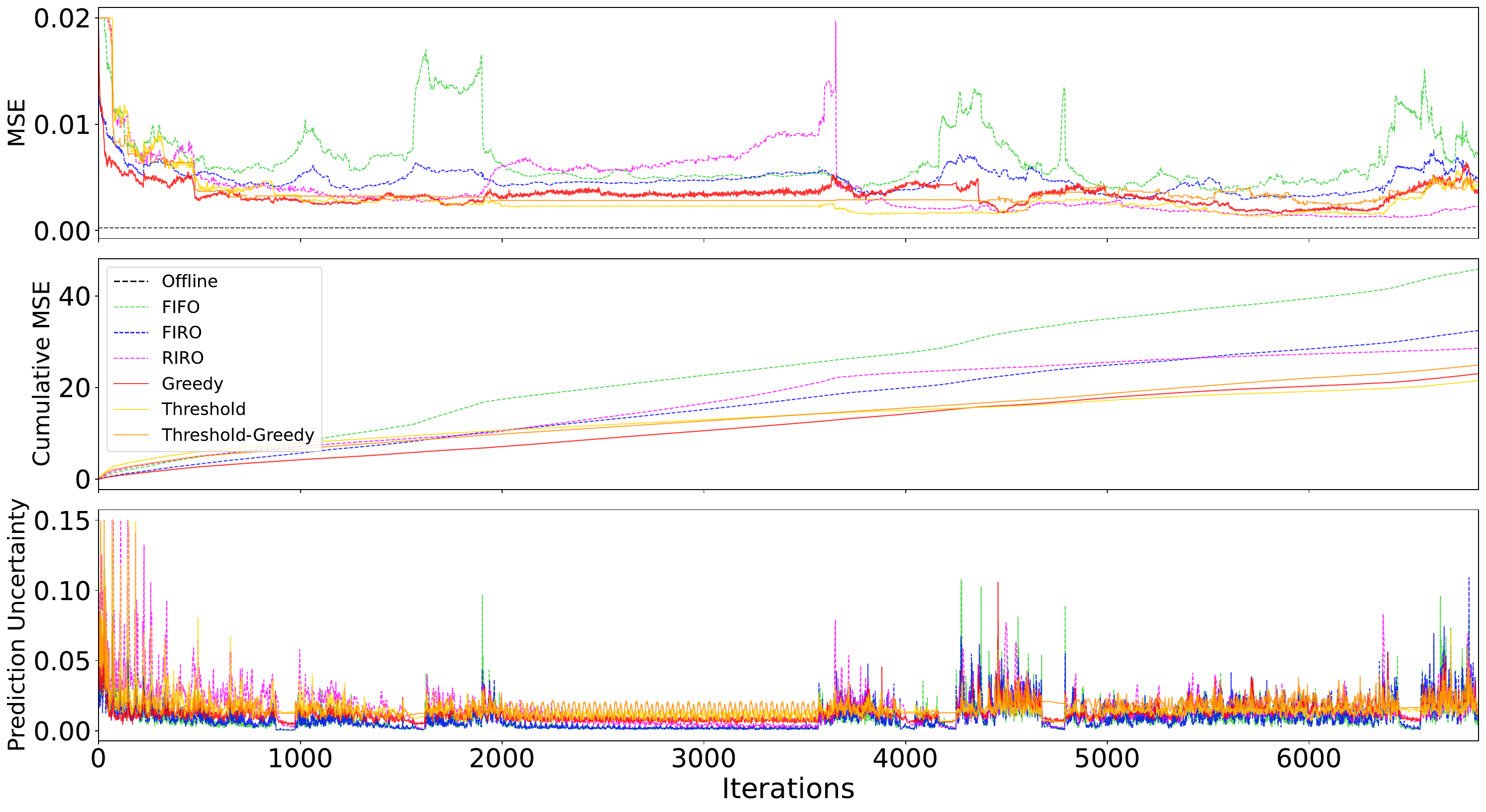}
    \caption{Final comparison of online learning models using Ensembles for UQ. \textmd{The x-axis denotes the iterations over the training set, and the y-axis the performance of respective metrics on the testing set in each moment. The baseline models are denoted by dotted lines and uncertainty models by full lines. The plots of MSE, $R^2$, and predicted uncertainty were truncated to make the graphs more interpretable.}} %
    \label{fig:final_go}
\end{figure*}

Using the optimal parameters, we can compare the performance of all the models. We select $p$ for RIRO as 0.2, $t$ for Threshold to be 0.0156, and $t$ for Threshold-Greedy to be 0.0228. The comparison of the performance over the training set with the metrics recorded over the testing set is shown in Figure \ref{fig:final_go}. On the shared x-axis, we can see the iterations over the training set, and on the y-axis, we can see the metric value, which is calculated over the testing test at each iteration. Each model has its assigned color, which is the same on all plots. The results of MSE and $R^2$ were truncated in the y-axis to remove outlying performance at the beginning of the learning process where the model performed poorly, which made the graphs not interpretable.

\subsection{Buffer Saturation}

Lastly, we look into the performance of the methods with a varying size of a buffer. For all of the previous experiments, the buffer size was set to 100 for all of the methods. To produce these results the optimal parameters were used, that is the same set of parameters which was used to produce the results shown in Figure \ref{fig:final_go}.  

\begin{figure}[t] 
    \centering
    \subfigure[Ensembles]{
        \includegraphics[width=\linewidth]{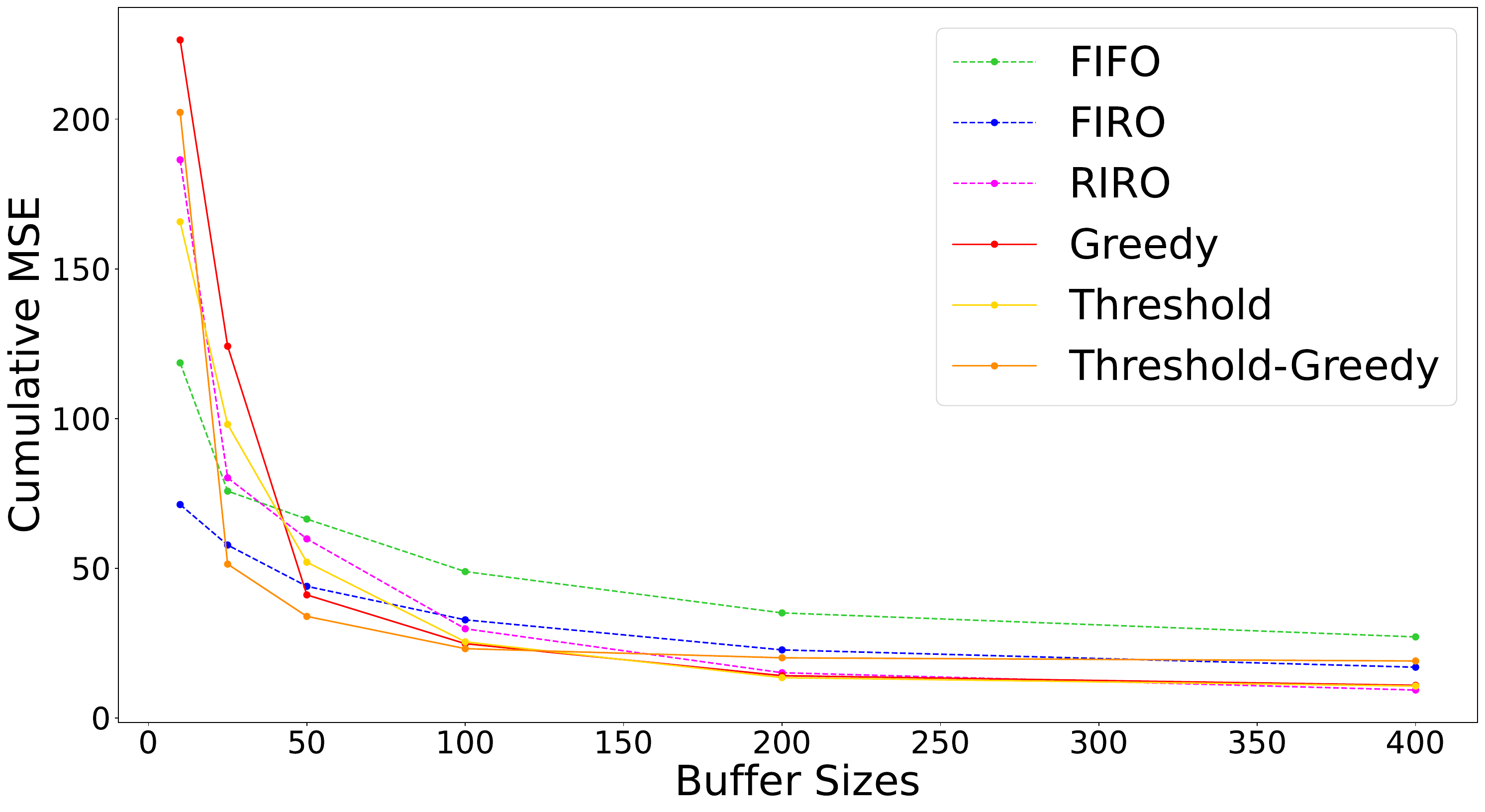}
    }
    \subfigure[Dropout]{
        \includegraphics[width=\linewidth]{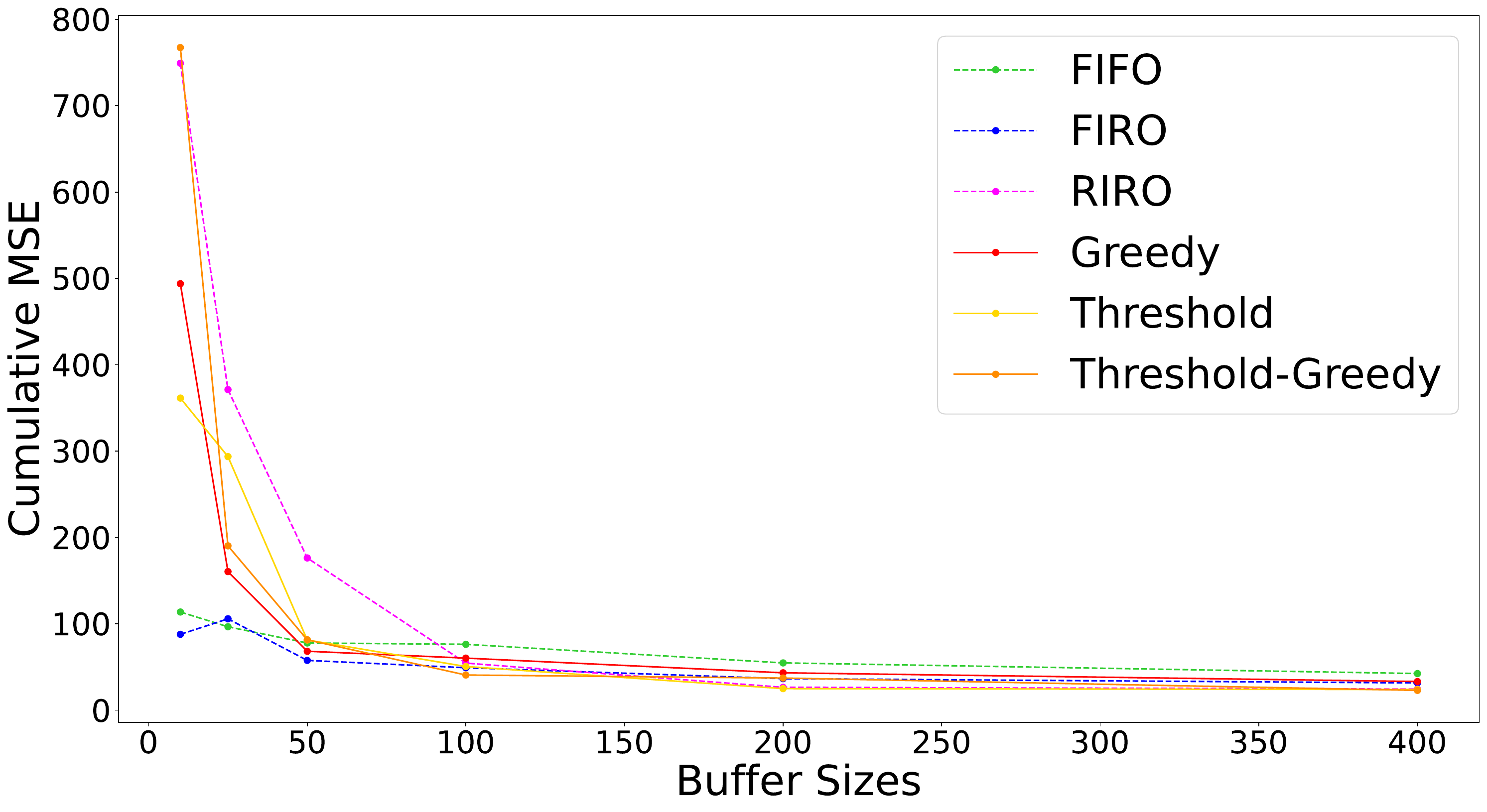}
    }
    \caption{Cumulative MSE of each method depending on the buffer size with Dropout and Ensembles UQ. \textmd{The buffer size influences the maximum number of points each method can store at any point during the online learning process. This has an influence on the model quality as the more points the model can store, the better the stored points represent the whole data distribution. The graph shows the final cumulative MSE after the learning process as a function of storage buffer size.}}
    \label{fig:buffer_scatter}
\end{figure}

The buffer size we choose to investigate is from the set $\{10, 20, 50, 100, 200, 400\}$. We look at the cumulative MSE achieved by each of the methods over the learning process given the size of the buffer. The results are presented in Figure  \ref{fig:buffer_scatter}.

\subsection{Method Comparison}

We wanted to repeat all the experiments with Dropout and Flipouts. We have used the same basic machine learning hyperparameters. We noticed Flipout was behaving strangely in the training, therefore we decided to further tune it, the results presented are achieved with the default $p=0.7$ and $t = 0.02$. Optimal parameters found for Dropout were $t = 0.018$ for both Threshold-Greedy and Threshold and $p = 0.3$ for RIRO. The results are shown in Table \ref{tab:consolidated_results}.

\begin{table*}[t]
    \centering
    \setlength\tabcolsep{4pt} 
    \begin{tabular}{lllllllll}
        \toprule
        Method & Metric & D & FIFO & FIRO & RIRO & Greedy & Threshold & Threshold-Greedy \\
        \midrule
        \multirow{4}{*}{Ensembles}  & Minimum MSE & $\downarrow$ & \textbf{0.0036} & \textbf{0.0029} & \textbf{0.0012} & \textbf{0.0016} & \textbf{0.0013} & \textbf{0.0023} \\
        & Mean $R^2$ & $\uparrow$ & \textbf{0.06} & \textbf{0.37} & \textbf{0.52} & \textbf{0.63} & \textbf{0.55} & \textbf{0.52} \\
        & Cumulative MSE & $\downarrow$ & \textbf{45.87} & \textbf{32.44} & \textbf{28.61} & \textbf{23.01} & \textbf{21.50} & \textbf{24.93} \\
        & \% Dataset Use & $\downarrow$ & NA & NA & \textbf{20.5\%} & 97.4\% & 28.5\% & 9.3\% \\
        \midrule
        \multirow{4}{*}{Dropout} & Minimum MSE & $\downarrow$ & 0.0053 & 0.0042 & 0.0033 & 0.0035 & 0.0030 & 0.0036 \\
        & Mean $R^2$ & $\uparrow$ & -0.29 & 0.10 & -0.10 & -0.17 & -0.52 & -0.05 \\
        & Cumulative MSE  & $\downarrow$ & 70.28 & 50.99 & 40.63 & 54.78 & 46.04 & 42.82 \\
        & \% Dataset Use & $\downarrow$ & NA & NA & 30.4\% & \textbf{81.0\%} & \textbf{10.8\%} & \textbf{7.4\%} \\
        \midrule
        \multirow{4}{*}{Flipout} & Minimum MSE & $\downarrow$ & 0.0350 & 0.0251 & 0.0524 & 0.0356 & 0.0544 & 0.0556 \\
        & Mean $R^2$ & $\uparrow$ & -0.12 & 0.13 & -0.21 & 0.00 & -0.37 & -0.66\\
        & Cumulative MSE & $\downarrow$ & 387.81 & 291.74 & 406.27 & 347.49 & 446.12 & 520.70 \\
        & \% Dataset Use & $\downarrow$ & NA & NA & 69.2\% & 98.9\% & 100\% & 99.9\% \\
        \bottomrule
    \end{tabular}
    \caption{Comparison of several metrics across all uncertainty and continual learning methods, indicating the direction (D) of improvement for each metric (lower or higher in comparison with other methods).}
    \label{tab:consolidated_results}
\end{table*}

\section{Discussion}\label{sec:discussion}

The aim of this paper was to determine, whether the amount of data can be retained in the online learning process without a loss of performance, thanks to uncertainty. Now we will draw a conclusion based on the results of performed experiments.

\textbf{Parameter Tuning}. The performance of models with different parameters $p$ and $t$ was compared based on the final cumulative MSE in Figure \ref{fig:Extra_Parameters_Tuning}. From the results, we can clearly conclude that lower $p$ is more beneficial to final cumulative loss, however, but as additional analysis of results shows, these models have higher MSE at the beginning. This exactly follows what we predicted in Section \ref{sec:proposed-methods}, as lower $p$ means that we need a longer time to establish an even distribution of the data over the whole sample space. Similarly to this finding, Figure \ref{fig:Extra_Parameters_Tuning} also suggests that higher thresholds guide models to higher performance on the final cumulative MSE.

\textbf{Optimal Parameters}. The results with the optimally tuned $p$ and $t$ are shown in Figure \ref{fig:final_go}. The graphs of MSE show that baselines (FIFO, FIRO, RIRO) do not obtain a stable model over the whole learning process. The MSE oscillates a lot, which indicates that these methods locally overfit the current data and do not manage to generalize over the whole learning process. They suffer from catastrophic forgetting constantly. This projects into the cumulative MSE metric, which tells us that overall FIFO performed the worst. FIRO and RIRO perform quite a bit better, but still worse than the uncertainty-based methods.

The uncertainty-based methods all manage to regularize for local overfitting on the current data. This is proven by the stability of the learning process, as the curves of MSE for Threshold, Greedy, and Threshold-Greedy all stay low during the whole learning process. Greedy converges the most quickly at the beginning of the training. However, later it is less stable. The Threshold method converges a lot slower at the beginning, with Threshold-Greedy performing somewhere in between the two. Later in the process, the Threshold method is the most stable and keeps to a good model, while other methods oscillate a lot more. All of these effects are captured in the cumulative MSE, which sets the Threshold to be the best-performing model overall, followed by Greedy and Threshold-Greedy. We note that the optimal value of $t$ for Threshold-Greedy was at the bound of the interval, which might have negatively influenced the performance.

However, the cumulative loss does not tell the full story. By looking at the development of MSE of RIRO at the end of the process, we can see it is the lowest of all methods. We can conclude that over a long period, taking just random points introduces sufficient robustness against local overfitting. However, it is important to note that RIRO was not as stable during the whole learning process, introducing a sharp decrease in performance in the middle of the learning process. From this we can conclude, that uncertainty is a good selection metric for picking the points, which can help us increase the robustness of the model, yet over long horizons, it may perform no better than random selection.

In the graph of prediction uncertainty we can see that statistical baseline methods predict lower uncertainty most of the time, with high peaks from time to time. This can be interpreted as overfitting the local data, as when very novel data appears, the uncertainty spikes. In contrast, the uncertainty-based models, predict somewhat higher uncertainty over most of the points, however, they do not have such high peaks. This indicates that these methods manage to generalize and novel data is only a slightly bit uncertain.

It is interesting to consider the performance of all the models in comparison with the Offline baseline, which has considered all the points at once. The performance of the baseline has been plotted as a line for MSE over the whole process. We conclude that no one of the models came close to achieving the same MSE as the baseline. RIRO was closest at the end, achieving an MSE of 0.001236, however, the Offline baseline obtained an MSE equal to 0.000237. From this, we can conclude, that the buffer of size 100 is simply not big enough to contain sufficient data for learning models that can achieve a minimal loss.

\textbf{Buffer Size Influence}. As hypothesized, the performance in terms of cumulative loss changes with the size of the buffer, as shown in Figure \ref{fig:buffer_scatter}. For all of the models, the bigger the buffer, the smaller the final cumulative loss. The gain related to the increase in the size of the buffer is not the same for all the methods. All in all, the uncertainty-based methods perform poorly with a small buffer size, while for the bigger buffer sizes, the performance is getting better and better. The baseline methods, on the other hand, show a more stable performance regardless of the size of the buffer. For all of the methods, the gain in performance is indeed diminishing with the increased buffer size.

\textbf{UQ Comparison}. Table \ref{tab:consolidated_results} summarizes the comparison results. Selection of Ensembles was made for the first methods, as they are the best in the estimation of epistemic uncertainty \cite{valdenegro2022deeper}. Ensembles outperform Dropout in all metrics. However, we can still conclude that Dropout works decently well for the task. We also need to consider that Dropout is less computationally heavy, which might come as an advantage. The results of Flipout are less promising, as it performs worse than the baseline methods in all metrics. We hypothesize that this is due to instability of training of the network. Also as shown in \cite{valdenegro2022deeper}, Flipout does not show good capability in modelling epistemic uncertainty.

\section{Conclusions and Future Work}

Based on the results we presented, we conclude that the uncertainty-based methods exhibit greater stability over the learning process and consume fewer data points than statistical baselines. All in all, the Threshold method has performed the most stable and delivered the best performance of all models. Additionally, we can conclude that the epistemic uncertainty quantification quality of the model is crucial to the task. This sets the simple ensemble to be the most useful method for the online learning of AUV model dynamics.

As noted in Table \ref{tab:consolidated_results}, the method that required the least data was the Threshold-Greedy method, which used only 9.3\% of the data and performed only a bit worse than the Threshold method. This cut in used resources has to be put into perspective: in the process of uncertainty estimation, 10 models in the ensemble were used, which all had to be trained separately. So all in all, we conclude that there is no free lunch \cite{adam2019no}. One might consider using the Dropout method instead, as it is less computationally demanding and sacrifices a bit of performance.

We expect that our proposed methods and results would allow for Autonomous Underwater Vehicles to perform online learning of their dynamics models and profit from uncertainty estimation to reduce computational training loads and data storage requirements. Both can increase the long-term autonomy of AUVs.

There are potential flaws in the study, which could be improved upon. One might argue that fitting the parameters separately in two rounds (firstly the basic machine learning parameters, then the other parameters of the methods) is not fully fair. The parameter for the Threshold that was guessed for the initial hyperparameter tuning was closer to the optimal one than the one chosen for RIRO. This could have negatively influenced the final performance of the RIRO method. It would be optimal, to tune all the parameters at once, refine the search space, and give the Bayesian Optimizer more iterations.

Although we have performed an extensive comparison of uncertainty-based methods against statistical baselines, it would be useful to compare the results also with some spatial and temporal heuristic approaches, as presented in \cite{wehbe2017online}. Achieving equally-spaced spatial distribution of points in the dataset might incur smaller computational costs while inducing the great performance of the model. It would be useful to see, how the uncertainty-based methods would compare against such approaches.

Although we have concluded that the threshold method performed the best, the performance of the methods is closely tied to the parameter $t$. This means, that the parameter must be tuned up front, which might not be very useful in practice. It would be great to investigate an adaptive approach, for example,  using percentiles. Let the model estimate the uncertainties and only accept those in the 90\% percentile of uncertainty. Such a technique would remove dependence from the tuning procedure.

All in all, the ultimate proof of the usefulness of the suggested methods will be the deployment of the methods in the real environment on an autonomous underwater vehicle. This would be the next suggested step towards seeing how the method could be exploited in practice.

\section{Acknowledgements}
We thank the Center for Information Technology of the University of Groningen for their support and for providing access to the Hábrók high-performance computing cluster.
\bibliographystyle{IEEEtran}
\bibliography{literature}

\begin{thebibliography}{10}
\providecommand{\url}[1]{#1}
\csname url@rmstyle\endcsname
\providecommand{\newblock}{\relax}
\providecommand{\bibinfo}[2]{#2}
\providecommand\BIBentrySTDinterwordspacing{\spaceskip=0pt\relax}
\providecommand\BIBentryALTinterwordstretchfactor{4}
\providecommand\BIBentryALTinterwordspacing{\spaceskip=\fontdimen2\font plus
\BIBentryALTinterwordstretchfactor\fontdimen3\font minus
  \fontdimen4\font\relax}
\providecommand\BIBforeignlanguage[2]{{%
\expandafter\ifx\csname l@#1\endcsname\relax
\typeout{** WARNING: IEEEtran.bst: No hyphenation pattern has been}%
\typeout{** loaded for the language `#1'. Using the pattern for}%
\typeout{** the default language instead.}%
\else
\language=\csname l@#1\endcsname
\fi
#2}}

\bibitem{wehbe2019framework}
B.~Wehbe, M.~Hildebrandt, and F.~Kirchner, ``A framework for on-line learning
  of underwater vehicles dynamic models,'' in \emph{2019 International
  Conference on Robotics and Automation (ICRA)}.\hskip 1em plus 0.5em minus
  0.4em\relax IEEE, 2019, pp. 7969--7975.

\bibitem{thuruthel2017learning}
T.~G. Thuruthel, E.~Falotico, F.~Renda, and C.~Laschi, ``Learning dynamic
  models for open loop predictive control of soft robotic manipulators,''
  \emph{Bioinspiration \& biomimetics}, vol.~12, no.~6, p. 066003, 2017.

\bibitem{wehbe2017learning}
B.~Wehbe and M.~M. Krell, ``Learning coupled dynamic models of underwater
  vehicles using support vector regression,'' in \emph{OCEANS
  2017-Aberdeen}.\hskip 1em plus 0.5em minus 0.4em\relax IEEE, 2017, pp. 1--7.

\bibitem{henson1998nonlinear}
M.~A. Henson, ``Nonlinear model predictive control: current status and future
  directions,'' \emph{Computers \& Chemical Engineering}, vol.~23, no.~2, pp.
  187--202, 1998.

\bibitem{wehbe2020long}
B.~Wehbe, ``Long-term adaptive modeling for autonomous underwater vehicles,''
  Ph.D. dissertation, Universit{\"a}t Bremen, 2020.

\bibitem{french1999catastrophic}
R.~M. French, ``Catastrophic forgetting in connectionist networks,''
  \emph{Trends in cognitive sciences}, vol.~3, no.~4, pp. 128--135, 1999.

\bibitem{verwimp2021rehearsal}
E.~Verwimp, M.~De~Lange, and T.~Tuytelaars, ``Rehearsal revealed: The limits
  and merits of revisiting samples in continual learning,'' in
  \emph{Proceedings of the IEEE/CVF International Conference on Computer
  Vision}, 2021, pp. 9385--9394.

\bibitem{8329822}
J.~L. Part and O.~Lemon, ``Incremental online learning of objects for robots
  operating in real environments,'' in \emph{2017 Joint IEEE International
  Conference on Development and Learning and Epigenetic Robotics
  (ICDL-EpiRob)}, 2017, pp. 304--310.

\bibitem{settles2009active}
B.~Settles, ``Active learning literature survey,'' \emph{MINDS@UW}, 2009.

\bibitem{valdenegro2022deeper}
M.~Valdenegro-Toro and D.~S. Mori, ``A deeper look into aleatoric and epistemic
  uncertainty disentanglement,'' in \emph{2022 IEEE/CVF Conference on Computer
  Vision and Pattern Recognition Workshops (CVPRW)}.\hskip 1em plus 0.5em minus
  0.4em\relax IEEE, 2022, pp. 1508--1516.

\bibitem{cantelobre2020real}
T.~Cantelobre, C.~Chahbazian, A.~Croux, and S.~Bonnabel, ``A real-time
  unscented kalman filter on manifolds for challenging auv navigation,'' in
  \emph{2020 IEEE/RSJ International Conference on Intelligent Robots and
  Systems (IROS)}.\hskip 1em plus 0.5em minus 0.4em\relax IEEE, 2020, pp.
  2309--2316.

\bibitem{paine2018adaptive}
T.~M. Paine and L.~L. Whitcomb, ``Adaptive parameter identification of
  underactuated unmanned underwater vehicles: A preliminary simulation study,''
  in \emph{OCEANS 2018 MTS/IEEE Charleston}.\hskip 1em plus 0.5em minus
  0.4em\relax IEEE, 2018, pp. 1--6.

\bibitem{gibson2018hydrodynamic}
S.~B. Gibson and D.~J. Stilwell, ``Hydrodynamic parameter estimation for
  autonomous underwater vehicles,'' \emph{IEEE Journal of Oceanic Engineering},
  vol.~45, no.~2, pp. 385--394, 2018.

\bibitem{harris2023stable}
Z.~J. Harris, A.~M. Mao, T.~M. Paine, and L.~L. Whitcomb, ``Stable nullspace
  adaptive parameter identification of 6 degree-of-freedom plant and actuator
  models for underactuated vehicles: Theory and experimental evaluation,''
  \emph{The International Journal of Robotics Research}, vol.~42, no.~12, pp.
  1070--1093, 2023.

\bibitem{ramirez2021dynamic}
W.~A. Ramirez, J.~Kocijan, Z.~Q. Leong, H.~D. Nguyen, and S.~G. Jayasinghe,
  ``Dynamic system identification of underwater vehicles using multi-output
  gaussian processes,'' \emph{International Journal of Automation and
  Computing}, vol.~18, no.~5, pp. 681--693, 2021.

\bibitem{bande2021online}
M.~Bande and B.~Wehbe, ``Online model adaptation of autonomous underwater
  vehicles with lstm networks,'' in \emph{OCEANS 2021: San Diego--Porto}.\hskip
  1em plus 0.5em minus 0.4em\relax IEEE, 2021, pp. 1--6.

\bibitem{cohn1996active}
D.~A. Cohn, Z.~Ghahramani, and M.~I. Jordan, ``Active learning with statistical
  models,'' \emph{Journal of artificial intelligence research}, vol.~4, pp.
  129--145, 1996.

\bibitem{dasgupta2004analysis}
S.~Dasgupta, ``Analysis of a greedy active learning strategy,'' \emph{Advances
  in neural information processing systems}, vol.~17, 2004.

\bibitem{williams1995gaussian}
C.~Williams and C.~Rasmussen, ``Gaussian processes for regression,''
  \emph{Advances in neural information processing systems}, vol.~8, 1995.

\bibitem{scholkopf2000new}
B.~Sch{\"o}lkopf, A.~J. Smola, R.~C. Williamson, and P.~L. Bartlett, ``New
  support vector algorithms,'' \emph{Neural computation}, vol.~12, no.~5, pp.
  1207--1245, 2000.

\bibitem{ghahramani1999variational}
Z.~Ghahramani and M.~Beal, ``Variational inference for bayesian mixtures of
  factor analysers,'' \emph{Advances in neural information processing systems},
  vol.~12, 1999.

\bibitem{vijayakumar2005incremental}
S.~Vijayakumar, A.~D'souza, and S.~Schaal, ``Incremental online learning in
  high dimensions,'' \emph{Neural computation}, vol.~17, no.~12, pp.
  2602--2634, 2005.

\bibitem{sculley2007online}
D.~Sculley, ``Online active learning methods for fast label-efficient spam
  filtering.'' in \emph{CEAS}, vol.~7, 2007, p. 143.

\bibitem{he2020incremental}
J.~He, R.~Mao, Z.~Shao, and F.~Zhu, ``Incremental learning in online
  scenario,'' in \emph{Proceedings of the IEEE/CVF conference on computer
  vision and pattern recognition}, 2020, pp. 13\,926--13\,935.

\bibitem{chen2022online}
C.~Chen, Y.~Li, and Y.~Sun, ``Online active regression,'' in
  \emph{International Conference on Machine Learning}.\hskip 1em plus 0.5em
  minus 0.4em\relax PMLR, 2022, pp. 3320--3335.

\bibitem{loeffel2017adaptive}
P.-X. Loeffel, ``Adaptive machine learning algorithms for data streams subject
  to concept drifts,'' Ph.D. dissertation, Universit{\'e} Pierre et Marie
  Curie-Paris VI, 2017.

\bibitem{paine2023ensemble}
T.~M. Paine and M.~R. Benjamin, ``An ensemble of online estimation methods for
  one degree-of-freedom models of unmanned surface vehicles: applied theory and
  preliminary field results with eight vehicles,'' in \emph{2023 IEEE/RSJ
  International Conference on Intelligent Robots and Systems (IROS)}.\hskip 1em
  plus 0.5em minus 0.4em\relax IEEE, 2023, pp. 6177--6184.

\bibitem{topini2020lstm}
E.~Topini, A.~Topini, M.~Franchi, A.~Bucci, N.~Secciani, A.~Ridolfi, and
  B.~Allotta, ``Lstm-based dead reckoning navigation for autonomous underwater
  vehicles,'' in \emph{Global Oceans 2020: Singapore--US Gulf Coast}.\hskip 1em
  plus 0.5em minus 0.4em\relax IEEE, 2020, pp. 1--7.

\bibitem{wang2021efficient}
S.~Wang, C.~Deng, and Q.~Qi, ``Efficient online calibration for autonomous
  vehicle’s longitudinal dynamical system: A gaussian model approach,'' in
  \emph{2021 IEEE International Conference on Robotics and Automation
  (ICRA)}.\hskip 1em plus 0.5em minus 0.4em\relax IEEE, 2021, pp. 5410--5416.

\bibitem{hildebrandt2010design}
M.~Hildebrandt and J.~Hilljegerdes, ``Design of a versatile auv for high
  precision visual mapping and algorithm evaluation,'' in \emph{2010 IEEE/OES
  Autonomous Underwater Vehicles}.\hskip 1em plus 0.5em minus 0.4em\relax IEEE,
  2010, pp. 1--6.

\bibitem{valdenegro2021exploring}
M.~Valdenegro-Toro, ``Exploring the limits of epistemic uncertainty
  quantification in low-shot settings,'' \emph{arXiv preprint
  arXiv:2111.09808}, 2021.

\bibitem{gal2016dropout}
Y.~Gal and Z.~Ghahramani, ``Dropout as a bayesian approximation: Representing
  model uncertainty in deep learning,'' in \emph{international conference on
  machine learning}.\hskip 1em plus 0.5em minus 0.4em\relax PMLR, 2016, pp.
  1050--1059.

\bibitem{wen2018flipout}
Y.~Wen, P.~Vicol, J.~Ba, D.~Tran, and R.~Grosse, ``Flipout: Efficient
  pseudo-independent weight perturbations on mini-batches,'' \emph{arXiv
  preprint arXiv:1803.04386}, 2018.

\bibitem{lakshminarayanan2017simple}
B.~Lakshminarayanan, A.~Pritzel, and C.~Blundell, ``Simple and scalable
  predictive uncertainty estimation using deep ensembles,'' \emph{Advances in
  neural information processing systems}, vol.~30, 2017.

\bibitem{adam2019no}
S.~P. Adam, S.-A.~N. Alexandropoulos, P.~M. Pardalos, and M.~N. Vrahatis, ``No
  free lunch theorem: A review,'' \emph{Approximation and optimization:
  Algorithms, complexity and applications}, pp. 57--82, 2019.

\bibitem{wehbe2017online}
B.~Wehbe, A.~Fabisch, and M.~M. Krell, ``Online model identification for
  underwater vehicles through incremental support vector regression,'' in
  \emph{2017 IEEE/RSJ International Conference on Intelligent Robots and
  Systems (IROS)}.\hskip 1em plus 0.5em minus 0.4em\relax IEEE, 2017, pp.
  4173--4180.

\end{thebibliography}

\clearpage
\begin{appendices}

\twocolumn[\section{Visualization of Applied Methods}\label{ap:appendix-gifs}

The following figures visualize the methods used during the experiments using an example of a simple toy task. The dataset is composed of the values of the functions $\sin$ over a short range. During the incremental learning task, they are served to the model in order with increasing $x$. The model learns the function and visualizes its uncertainty over the whole interval. The visualization was created using an ensemble of 10 models. The following figures are gifs that can be viewed using pdf reader with a Javascript extension, for example, \href{https://okular.kde.org/}{Okular} or \href{https://get.adobe.com/reader/}{Adobe Reader}. Furthermore, for the gifs to be activated, the current page has to be selected. This can be ensured by viewing the document in 1-page mode and viewing this single page at once.

]

\begin{figure*}[b]
    \vspace{-20cm}

    \centering
    
    \subfigure[FIFO]
    {
        \animategraphics[loop,autoplay,width=0.43\linewidth]{10}{img/gifs/FIFO/iteration-}{0}{49}
        \label{gif:fifo}
    }
    \subfigure[FIRO]
    {
        \animategraphics[loop,autoplay,width=0.43\linewidth]{10}{img/gifs/FIRO/iteration-}{0}{49}

        \label{gif:firo}
    }
    \\[-2ex]
    \subfigure[RIRO]
    {
        \animategraphics[loop,autoplay,width=0.43\linewidth]{10}{img/gifs/RIRO/iteration-}{1}{33}
        \label{gif:riro}
    }
    \subfigure[Greedy]
    {
        \animategraphics[loop,autoplay,width=0.43\linewidth]{10}{img/gifs/GREEDY/iteration-}{0}{49}
        \label{gif:greedy}
    }
    \\[-2ex]
    \subfigure[Threshold]
    {
        \animategraphics[loop,autoplay,width=0.43\linewidth]{10}{img/gifs/THRESHOLD/iteration-}{0}{41}
        \label{gif:threshold}
    }
    \subfigure[Threshold-Greedy]
    {
        \animategraphics[loop,autoplay,width=0.43\linewidth]{10}{img/gifs/THRESHOLD_GREEDY/iteration-}{0}{44}
        \label{gif:threshold-greedy}
    }
    \caption{Animated visualization of data selection methods treated in the paper.}
    \label{gif:all_figs}
\end{figure*}

\end{appendices}

\end{document}